# Vision and Contact based Optimal Control for Autonomous Trocar Docking

Christopher E. Mower[1], Martin Huber[1], Huanyu Tian[1,3], Ayoob Davoodi[2],
Emmanuel Vander Poorten[2], Tom Vercauteren[1], and Christos Bergeles[1]

[1]*School of Biomedical Engineering & Imaging Sciences, King's College London, UK*
[2]*Department of Mechanical Engineering, KU Leuven, Leuven, Belgium*
[3]*Beijing Institute of Technology, Beijing, China*

**INTRODUCTION**

Future operating theaters will be equipped with robots to perform various surgical tasks including, for example, endoscope control (Fig. 1b). Fully automated surgery is likely to be part of the distant future [1]. However, human-in-the-loop supervisory control architectures where the surgeon selects from several autonomous sequences is already being successfully applied in preclinical tests, e.g. autonomous bowel anastomoses was recently carried out in animals [2]. Inserting an endoscope into a trocar or introducer is a key step for every keyhole surgical procedure – hereafter we will only refer to this device as a "trocar". Our goal is to develop a controller for autonomous trocar docking.

Autonomous trocar docking is a version of the *peg-in-hole* problem. Extensive work in the robotics literature addresses this problem, e.g. [3], [4], [5]. The peg-in-hole problem has been widely studied in the context of assembly [6] where, typically, the hole is considered static and rigid to interaction. In our case, however, the trocar is not fixed and responds to interaction. Within the scope of surgical robotics, a recent article addressed autonomous trocar docking for retinal surgery [7]. A vision-based system tracks the trocar and aligns the robot with the estimated pose prior to insertion. This work assumes the robot accurately aligns with the estimated pose, prior to insertion, and does not utilize contact information. In contrast, we consider procedures on a larger scale, e.g. endoscopic lumbar discectomy/decompression, and cholecystectomy. For these cases, often, surgeons will utilize contact between the endoscope and trocar in order to complete the insertion successfully. To the best of our knowledge, we have not found literature that explores this particular generalization of the problem directly.

Our primary contribution in this work is an optimal control formulation for automated trocar docking. We use a nonlinear optimization program to model the task, minimizing a cost function subject to constraints to find optimal joint configurations. The controller incorporates a geometric model for insertion and a force-feedback (FF) term to ensure patient safety by preventing excessive interaction forces with the trocar. Experiments, demonstrated on a real hardware lab setup, validate the approach (Fig. 1b). Our method successfully achieves trocar insertion on our real robot lab setup, and simulation trials demonstrate its ability to reduce interaction forces.

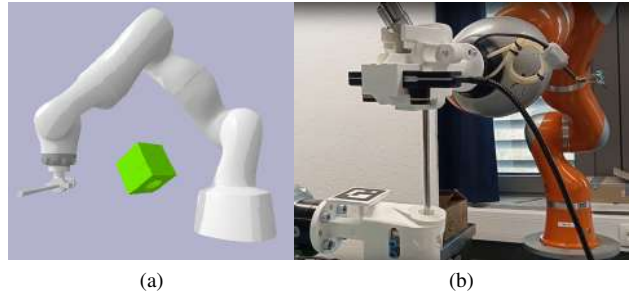

(a) (b)

Fig. 1: A robot arm equipped with an endoscope. (a) Robot simulation, the green box is a simulated trocar that responds to interaction, and (b) hardware realization.

**MATERIALS AND METHODS**

*Optimal Control Approach*

We formulate a nonlinear controller, given by

$$q^* = \arg\min_q \sum_i w_i c_i(q; \widehat{p}) \quad \text{subject to} \quad q \in \mathbb{Q} \quad (1)$$

where $q \in \mathbb{R}^n$ is the $n$ degree of freedom robot joint position, $0 \leq c_i(\cdot) \in \mathbb{R}$ is a cost term for a sub-task with a corresponding scalar weight $0 < w_i \in \mathbb{R}$, $\widehat{p} \in \mathbb{R}^{n_p}$ is an arbitrary array of parameters from sensing data (a hat indicates the value is measured by a sensor), e.g. trocar pose that is assumed known, and $\mathbb{Q}$ is the space of feasible joint states (i.e. joint position/velocity limits). Motion is generated by first stacking sensing data into $\widehat{p}$, then solving (1), that results in a reference configuration $q^*$ to which the robot is controlled.

The terms $c_i$ for $i = 1{:}5$ model the tasks: (i) move the end-effector tip $e(q)$ towards the trocar insertion axis given by the direction vector $g_a$, modeled by $c_1 = \|e(q) - n(g_a)\|^2$ ($n(\cdot)$ is the nearest point on the line $g_a$), (ii) move end-effector towards a goal position $g_p$, modeled by $c_2 = \|e(q) - g_p\|^2$, (iii) align the endoscope optical axis $a(q)$ with the given by $c_3 = \|a(q) - g_a\|^2$, (iv) minimize joint velocity modeled by $c_4 = \|q - q_{prev}\|^2$, and (v) use FF measured from the robot joint external torques to reduce interaction forces, modeled by $c_5 = \|e(q) - r\|$ such that $r$ is found by an admittance controller based on the external torques measured at the joints.

We implement (1) using the OpTaS library [8], and use an sequential optimization programming approach as the solver. It is possible to linearize the problem about the current robot configuration $\widehat{q}_c$, enabling us to use a quadratic programming solver (typically faster). However, we observed that the nonlinear solver was able to reliably

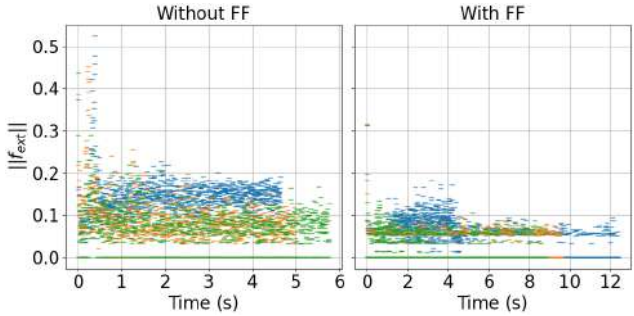

Fig. 2: Comparison of $\|f_{ext}\|$ (measured in Newtons) over time, from simulation trials, with/without FF. Color indicates a trial.

converge within 2ms - well within a sampling frequency amenable for online control.

*Simulation*

We employ the PyBullet [9] simulator for reliable contact modeling [10], a KUKA LBR Med robot arm with an endoscope attachment (Fig. 1a), and a trocar represented by a cloth-like mesh attached to a holed box. To simulate real-world conditions, we introduce (Gaussian) noise to the trocar pose retrieval.

*Hardware realization*

Our hardware setup, depicted in Fig. 1b, consists of a custom mount with an endoscope and a ZEDm (Stereolabs, USA) camera for state estimation attached to the robot end-effector. The endoscope tip position relative to the robot end-effector is determined through CAD model measurements, while the camera pose is estimated using eye-in-hand calibration (i.e. wiki.ros.org/handeye). Although the camera can capture depth images, a single RGB image stream is used. At this stage a trocar, a trocar with an attached ArUco marker is placed in a position visible to the camera. In the current setup, the trocar is positioned in a fixed location. The setup has successfully completed the insertion task in multiple trials. Future work will explore removing the visible marker requirement, as discussed later.

## RESULTS

In simulation, we compare with/without a FF term to indicate the method leads to reduced interactions with the trocar. The interaction forces $\widehat{f}_{ext} \in \mathbb{R}^3$ at the endoscope tip are estimated using $\widehat{f}_{ext} = J_{lin}(\widehat{q}_c)^{-T} \widehat{\tau}_{ext}$ where $J_{lin}(\cdot) \in \mathbb{R}^{3 \times 7}$ is the linear geometric Jacobian matrix, and $\widehat{\tau}_{ext} \in \mathbb{R}^7$ are external torques measured at the joints.

For three trials with/without FF, the distributions of $\|f_{ext}\|$ are shown in Fig. 2. Visually, it is apparent that the interaction forces are reduced when incorporating the FF into the controller. Note, the time required to complete the task is longer when incorporating FF. These results suggest a reduction in $\|f_{ext}\|$, that is quantified by the performance metric $M = \frac{1}{T} \int_0^T \|f_{ext}\| \, dt$, $T$ is the completion time. The metric $M$ is the normalized force integral; lower values indicate higher performance. For each trial we compute $M$ and take the average. The results without and with FF respectively is 21.4 ± 9.1N and 3.4 ± 1.6N.

## DISCUSSION

We successfully developed and implemented an autonomous trocar docking system in a realistic lab setup. Comparisons indicate that incorporating FF can improve safety by reducing interaction forces.

The hardware realization validates our pipeline but has limitations. State estimation relies on marker visibility, we intend to explore a state estimator trained on synthetic, markerless data. Our current controller focuses on linear interaction forces during the initial stage of the task, before insertion, as rotational forces are not significant at this stage. However, we acknowledge that rotational moments are expected, especially after partial trocar insertion. In future iterations, we will incorporate these to ensure compliance in the full 6D task space. Additionally, our next hardware setup will assume a compliant trocar.

We will also investigate kinesthetic teaching to eliminate the need for extensive parameter tuning. However, decoupling interaction forces between the trocar and the demonstrator might pose challenges. Teleoperation-based demonstrations are an alternative, but collecting sufficient training data could be laborious due to the small insertion port (approximately 7mm) and potential view-point issues.

## ACKNOWLEDGMENTS

This project has received funding from the European Union's Horizon 2020 research and innovation programme under grant agreement No 101016985 (FAROS project).